\pdfoutput=1

\documentclass[11pt]{article}

\usepackage{acl}

\usepackage{times}
\usepackage{latexsym}

\usepackage[T1]{fontenc}

\usepackage[utf8]{inputenc}

\usepackage{microtype}
\usepackage{booktabs}
\usepackage{graphicx}
\usepackage{soul}

\newcommand{\eg}{e.g.~}
\newcommand{\ie}{i.e.~}

\title{Estimating Numbers without Regression}

\author{Avijit Thawani \\
  Univ of Southern California \\
  \texttt{thawani@usc.edu} \\\And
  Jay Pujara \\
  Univ of Southern California \\
  Information Sciences Institute \\\And
  Ashwin Kalyan \\
  Allen Institute for \\
  Artificial Intelligence \\
 }

\begin{document}

\maketitle

\begin{abstract}
  Despite recent successes in language models, their ability to represent numbers is insufficient. Humans conceptualize numbers based on their magnitudes, effectively projecting them on a number line; whereas subword tokenization fails to explicitly capture magnitude by splitting numbers into arbitrary chunks.
  To alleviate this shortcoming, alternative approaches have been proposed that modify numbers at various stages of the language modeling pipeline. 
    These methods change either the (1) notation in which numbers are written (\eg scientific vs decimal), the (2) vocabulary used to represent numbers or the entire (3) architecture of the underlying language model, to directly regress to a desired number.

    Previous work \cite{berg-kirkpatrick-spokoyny-2020-empirical} suggests that architectural change helps achieve state-of-the-art on number estimation but we find an insightful ablation: changing the model's vocabulary instead (\eg introduce a new token for numbers in range 10-100) is a far better trade-off.
    In the context of masked number prediction, a carefully designed tokenization scheme is both the simplest to implement and sufficient, \ie with similar performance to the state-of-the-art approach that requires making significant architectural changes.
    Finally, we report similar trends on the downstream task of numerical fact estimation (for Fermi Problems) and discuss reasons behind our findings.
\end{abstract}

\section{Introduction}

The standard practice in the natural language processing (NLP) community is to process numbers in exactly the same manner as words.
This counter-intuitive treatment of numbers leads to their inaccurate representation and therefore, limited numerical understanding of large-scale language models (LMs) \citep{razeghi2022impact}.
To illustrate, a number like $\$799$ is \emph{subword} tokenized \citep{sennrich-etal-2016-neural} as $79$ and $\#\#9$.
Such a tokenization method, by construction, prevents accurately modeling the relationship of this number with others close on the number line say, $\$800$, as the surface forms share no common tokens. 

\begin{figure*}
    \centering
    \includegraphics[width=\textwidth]{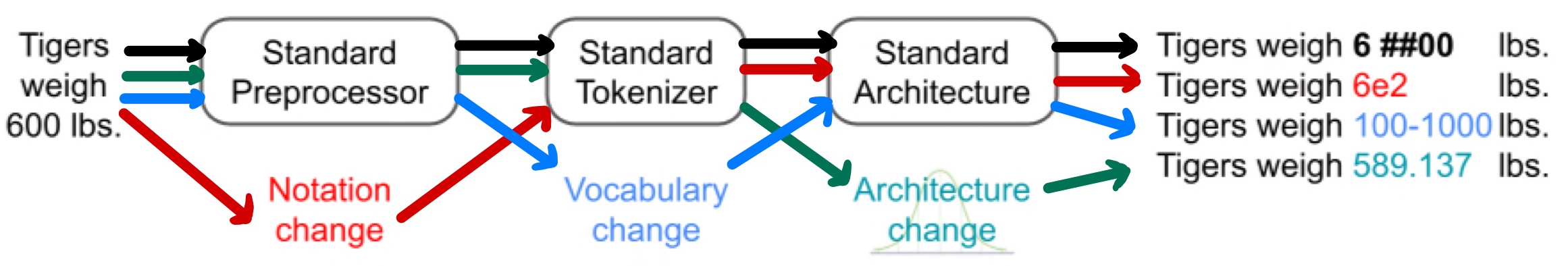}
    \caption{Alternative number representations change one of the three stages in the NLP pipeline.}
    \label{fig:method}
\end{figure*}

Many alternatives have been proposed to capture the scalar magnitude of numbers \citep{thawani-number-survey}.
All number decoders proposed to capture the magnitude of numbers fall into one of the following categories, corresponding to changes in 1) \textbf{notation} (\eg scientific vs decimal) or 2) \textbf{vocabulary} (\eg introducing new tokens that denote all numbers within a specified range)
or 3) \textbf{architectural} changes (\eg directly regressing to a number).
Figure \ref{fig:method} shows various alternative number representation methods ordered by increasing levels of intervention on a typical NLP pipeline, color coded consistently across the paper for legibility.

We find that applying the vocabulary-level changes leads to near state-of-the-art performance requiring no additional pretraining or architectural changes. 
This is a surprising yet useful ablation result, which can substantially speed up adoption of numeracy into any given language model. Any arbitrary LM can be made \emph{numerate} by simply tokenizing numbers on the number line.

We further evaluate the number representation schemes on their ability to generalize to a downstream task of numerical fact estimation in the context of solving Fermi problems \citep{fermi}.
We find similar trends, demonstrating the utility of the simple yet effective tokenization scheme in the decoding setting.
Finally, we discuss how these results may be explained by the observed distribution of mantissas in natural language.

\vspace{-5pt}
\section{Kinds of Number Representations}
\label{sec:methods}
\vspace{-5pt}

Our work focuses not on NLP models performing arithmetic, instead on their comprehensive understanding of approximate numbers, with the setting of masked number prediction (MNP) in natural language. 
This section introduces existing classes of number decoders and their respective trade-offs.

\paragraph{Subword.} 
The default way that language models decode numbers is the same way as words, one subword at a time, \eg, the number $600$ could be decoded as two individual tokens $6$ and $\#\#00$.

\paragraph{\textcolor{red}{Notation-change.}} 
Here, the numbers are represented in an alternative notation by preprocessing text before feeding into any off-the-shelf tokenizer and model.
We consider the following variations: 
\textbf{1. Scientific:} Using scientific notation, e.g., $6e2$ (where 6 is the mantissa and 2 is the exponent) in lieu of the usual decimal notation was first proposed by \citet{zhang2020scale}. In this work, we closely follow their version with minor implementation level changes. Note that following the notation change, the tokenizer nevertheless splits it into subwords. 
\textbf{2. Digits:} Here, the number is split into its constituent digits or characters, e.g., $600$ becomes $6$ $0$ $0$. This approach offers a consistent decomposition of numbers into digits as opposed to arbitrary subword segmentation, and has been proven effective on simple numeric probes as well as arithmetic word problems \cite{geva-etal-2020-injecting}.

\paragraph{\textcolor{blue}{Vocabulary change.}} 
Unlike words, the notion of distance or similarity is more obviously defined for numbers in terms of their separation on the number line, a cognitive tool that human beings are known to intuitively use to process numeracy \citep{dehaene2011number}. 
This forms the basis of a change of vocabulary: numbers within a specified range are collapsed into a single token (\eg 100-1000) -- at the cost of precise representation of numbers.
This approach does not modify the LM architecture, instead merely adds new tokens to the vocabulary.

\paragraph{\textcolor{teal}{Architecture change.}} 
Finally, several recent methods have modified the underlying language model to emit continuous values when predicting numbers. 
At their core, they operate by regressing to the desired number conditioned on the language context.
See \citet{berg-kirkpatrick-spokoyny-2020-empirical} for a thorough comparison within this class of methods. We directly compare against their best variant: Discrete Latent Exponents (DExp), which first models the exponent part of a number as a multinomial, then uses it to parameterize a truncated log normal distribution to sample the mantissa, a continuous value. Note that this is the highest level of intervention possible, thereby making the method ineffective whenever the underlying LM architecture is not accessible, say over an API. 


\vspace{-5pt}
\section{Experimental setup}
\label{sec:exp}
\vspace{-5pt}
\textbf{Task}: We evaluate the above decoders on the task of masked number prediction (MNP): Given a sentence with a mask (\eg \emph{``Tigers weigh [MASK] lbs."}), the model must predict a number as close as possible to the ground truth (\eg $600$). 

\begin{table*}[t!]
    \centering
    \small
    \begin{tabular}{l|rr|rr|rr}
        \toprule
         & \multicolumn{2}{c|}{FinNews} & \multicolumn{2}{c|}{FinNews-\$} & \multicolumn{2}{c}{SciDocs} \\
         \textbf{Metrics} & E-Acc $\uparrow$  & \multicolumn{1}{c|}{LogMAE$\downarrow$}  & E-Acc$\uparrow$ & \multicolumn{1}{c|}{LogMAE$\downarrow$}  & E-Acc$\uparrow$ &  \multicolumn{1}{c}{LogMAE$\downarrow$}  \\
         \midrule
         \multicolumn{2}{l}{\textbf{Baselines}} \\
         Train-Mean & $1.0 \pm 0.1\%$ & 7.69 & $6.0 \pm 0.4\%$ & 4.68 & $0.0 \pm 0.0\%$ & 8.81 \\
         Train-Median & $5.5 \pm 0.2\%$ & 1.88 & $10.6 \pm 0.5\%$ & 2.66 & $49.5 \pm 0.7\%$ & 0.83 \\
         Train-Mode & $24.2 \pm 0.4\%$ & 2.02 & $8.1 \pm 0.5\%$ & 6.30 & $49.5 \pm 0.7\%$ & 1.00 \\
         
         \midrule
         Subword-Pad8 & $63.6 \pm 0.5\%$ & 0.68 & $29.1 \pm 0.8\%$ & 1.36 & $68.0 \pm 0.6\%$ & 0.68 \\
        
         \midrule
         \multicolumn{2}{l}{\textcolor{red}{\textbf{Notation-change}}} & \\
         Digit-Pad17 & $52.2 \pm 0.5\%$ & 0.93 & $33.0 \pm 0.8\%$ & 1.37 & $55.1 \pm 0.5\%$ & 0.91 \\ 
         Scientific-Pad8 & $52.5 \pm 0.5\%$ & 0.84 & NA & NA & $71.1 \pm 0.6\%$ & 0.66 \\ 
          
         \midrule
         \multicolumn{2}{l}{\textcolor{blue}{\textbf{Vocabulary-change}}} & \\
         Vocab-AM & \hl{$74.4 \pm 0.4\% $} & 0.65 & \hl{$57.1 \pm 0.8\%$} & 0.93 & \hl{$81.2 \pm 0.5\%$} & 0.51 \\ 
         Vocab-GM & $73.7 \pm 0.4\% $ & \underline{0.60} & \hl{$57.0 \pm 0.8\%$} & \underline{0.92} & \hl{$81.3 \pm 0.5\%$} & \underline{0.44} \\ 
         \midrule
         \multicolumn{2}{l}{\textcolor{teal}{\textbf{Architecture-change}}} & \multicolumn{4}{c}{\citet{berg-kirkpatrick-spokoyny-2020-empirical}} \\
         DExp  & \hl{$74.6 \pm 0.4\% $} & \textbf{0.50} & \hl{$57.5 \pm 0.8\%$} & \textbf{0.89} & \hl{$81.2 \pm 0.5\%$} & \textbf{0.39} \\
         
         \bottomrule
    \end{tabular}
    \caption{Order of magnitude accuracy (E-Acc) and Log Mean Absolute Error (LogMAE) on test sets.
    }
    \label{tab:bert}
    \vspace{-5pt}
\end{table*} 

\paragraph{Datasets:}
We follow \citet{berg-kirkpatrick-spokoyny-2020-empirical} to finetune and evaluate our models on three datasets\footnote{Data URL: \url{https://github.com/dspoka/mnm}} -- Financial News Articles (FinNews),
its subset containing mostly price-based numbers (FinNews-\$), and Scientific Articles (SciDocs) \citep{Lo2020S2ORCTS}; all numbers in these datasets lie between $1$-$10^{16}$. 

\paragraph{Metrics:}
We evaluate using two metrics -- a) Exponent Accuracy (E-Acc) that checks whether the predicted answer is of the same order of magnitude as the ground truth and b) Log Mean Absolute Error (LogMAE). Confidence Intervals for Exponent Accuracy, a classification metric, are reported as the Wilson Score Interval \citep{wilson}: $a \pm z \sqrt{a(1-a)/n}$, where $a$ is the accuracy, $z$ is the constant ($2.58$ for $99$\% CI), and $n$ is the number of observations in the respective test set. 

\vspace{-5pt}

\paragraph{Baselines:}
Our primary baseline is the standard approach of subword tokenization. 
We require each number prediction to be 8 tokens long, with appropriate padding, to be able to fairly represent all numbers in our range.  
Additionally, we evaluate on three trivial baselines that make a constant prediction corresponding to the mean, median, and mode of all numbers in the training set.
\vspace{-5pt}

\paragraph{Models:} 
We compare against both notation-level changes \ie scientific and digit, with a padding of 8 and 17 respectively. 
Among the approaches that introduce architectural changes, we compare against the SotA method of DExp (see previous section).
Finally, we compare against two variations that introduce vocabulary level changes -- both discretize the number line with logarithmically sized bins (with base $10$). 
The two variants differ in how the mantissa is chosen -- the arithmetic mean ($5$) or the geometric mean ($\sqrt{10}$), named Vocab-AM and Vocab-GM, respectively.
\footnote{Note that Vocab-AM/GM are mere ablations to the DExp methods -- the regression head replaced by static mantissas.}

\vspace{-5pt}

\paragraph{Implementation:}
Following the setup in \citet{berg-kirkpatrick-spokoyny-2020-empirical}, our base language model is 12-layer BERT-base and we fine-tune all models with a batch-size of 32 for 10 epochs. We use early stopping with a patience of three on the validation loss. We use two learning rates 3e-5 and 1e-2 for all pretrained parameters and newly added parameters respectively.
Please see Appendix \ref{appendix} for more details.

\vspace{-5pt}

\section{Results}
\vspace{-5pt}

We \textbf{bold-face the best} and \underline{underline the next best} LogMAE scores in each column (dataset), and we \hl{highlight} exponent accuracies that are within 99\% confidence of the SotA E-Acc.
NA denotes subword models which were unable to emit valid numbers for at least 50\% of the examples. 

\begin{table*}[t!]
    \centering
    \small
    \begin{tabular}{l|rr|rr|rr}
    \toprule
    \textbf{Fermi-Real} & \multicolumn{2}{c|}{trained on FinNews} & \multicolumn{2}{c|}{trained on FinNews-\$} & \multicolumn{2}{c}{trained on SciDocs} \\
    {510 egs.} & E-Acc $\uparrow$  & \multicolumn{1}{c|}{LogMAE$\downarrow$}  & E-Acc$\uparrow$ & \multicolumn{1}{c|}{LogMAE$\downarrow$}  & E-Acc$\uparrow$ &  \multicolumn{1}{c}{LogMAE$\downarrow$}  \\
     \midrule
    Sub-Pad8 & $26 \pm 5\% $ & 2.38 & $16 \pm 4\% $ & 3.17 & \hl{$26 \pm 5\% $} & 2.84 \\
    Dig-Pad17 & $19 \pm 5\% $ & 2.58 & NA    & NA   & $23 \pm 5\% $ & 2.87 \\
    Sci-Pad8 & $25 \pm 5\% $ & 2.93 & NA    & NA   & $20 \pm 5\% $ & 2.75 \\
    Vocab-AM & \hl{$32 \pm 5\% $} & \underline{2.19} & \hl{$24 \pm 5\% $} & \textbf{2.42} & \hl{$27 \pm 5\% $} & \underline{2.42} \\
    DExp & \hl{$32 \pm 5\% $} & \textbf{2.13} & \hl{$25 \pm 5\% $} & \underline{2.51} & \hl{$28 \pm 5\% $} & \textbf{2.40}  \\
    
    \midrule
    \textbf{Fermi-Syn} & \multicolumn{2}{c|}{trained on FinNews} & \multicolumn{2}{c|}{trained on FinNews-\$} & \multicolumn{2}{c}{trained on SciDocs} \\
    {3437 egs.} & E-Acc $\uparrow$  & \multicolumn{1}{c|}{LogMAE$\downarrow$}  & E-Acc$\uparrow$ & \multicolumn{1}{c|}{LogMAE$\downarrow$}  & E-Acc$\uparrow$ &  \multicolumn{1}{c}{LogMAE$\downarrow$}  \\
    \midrule
    Sub-Pad8 & $29 \pm 2\% $ & 2.89 & $19 \pm 2\% $ & 3.25  & $39 \pm 2\% $ & 2.83 \\
    Dig-Pad17 & $23 \pm 2\% $ & 2.93 & NA    & NA    & $41 \pm 2\% $ & 2.87 \\
    Sci-Pad8 & $26 \pm 2\% $ & 3.06 & NA    & NA    & $27 \pm 2\% $ & 2.76 \\
    Vocab-AM & \hl{$39 \pm 2\% $} & \underline{2.61} & \hl{$41 \pm 2\% $} & \textbf{2.42} & \hl{$48 \pm 2\% $} & \underline{2.52} \\
    DExp & \hl{$39 \pm 2\% $} & \textbf{2.44} & \hl{$41 \pm 2\% $} & \underline{2.44} & \hl{$48 \pm 2\% $} & \textbf{2.48} \\
    \bottomrule
    \end{tabular}
    \caption{Downstream performance of main methods over fact estimation for solving Fermi Problems. 
    }
    \label{tab:down}
    \vspace{-5pt}
\end{table*}

\paragraph{Intrinsic results} (Table \ref{tab:bert})
\label{sec:results}
We find that the change of notation approaches are inferior to the subword baseline. 
This is in contrast to prior work on extrapolating the arithmetic abilities of language models by notation changes \citep{nogueira2021investigating,geva-etal-2020-injecting}. It suggests that simple pre-processing changes of notation are not sufficient for contextual understanding of numbers for language modeling.
Next, we find that the vocabulary change methods (Vocab-AM/GM) are at par or better than the architectural change model (DExp). 
The improvement from subword to the DExp model, is achievable (within statistical bounds) without modelling the mantissa at all!

\paragraph{Downstream transfer} (Table \ref{tab:down})
Given such trends in masked number prediction, we are interested in the utility of these models on a downstream number prediction task. 
For this purpose, we evaluate on numerical fact estimation using the Fermi Problems dataset \citep{fermi}\footnote{Data URL: \url{https://allenai.org/data/fermi}}, which consists of challenging estimation problems such as \emph{``How many tennis balls fit in a school bus?''} 
Solving such questions require estimating numeric facts \eg \emph{volume of tennis ball} \& \emph{length of bus}. 

We evaluate our models (trained with different number decoders on one of the three datasets) in a zero-shot setting on such annotated facts provided as part of both the real and synthetic datasets part of the Fermi problem dataset. 
The task setup is of masked number prediction as before, e.g., ``\emph{the size of a tennis ball is \texttt{[MASK]} cubic centimeters.}" 
We find similar trends as before \ie change of notation is insufficient while vocabulary-change approaches are equal or better than architectural changes -- highlighting that most of the gains could be retained by simply tokenizing in number space.

\begin{figure}
    \centering
    \includegraphics[width=0.45\textwidth]{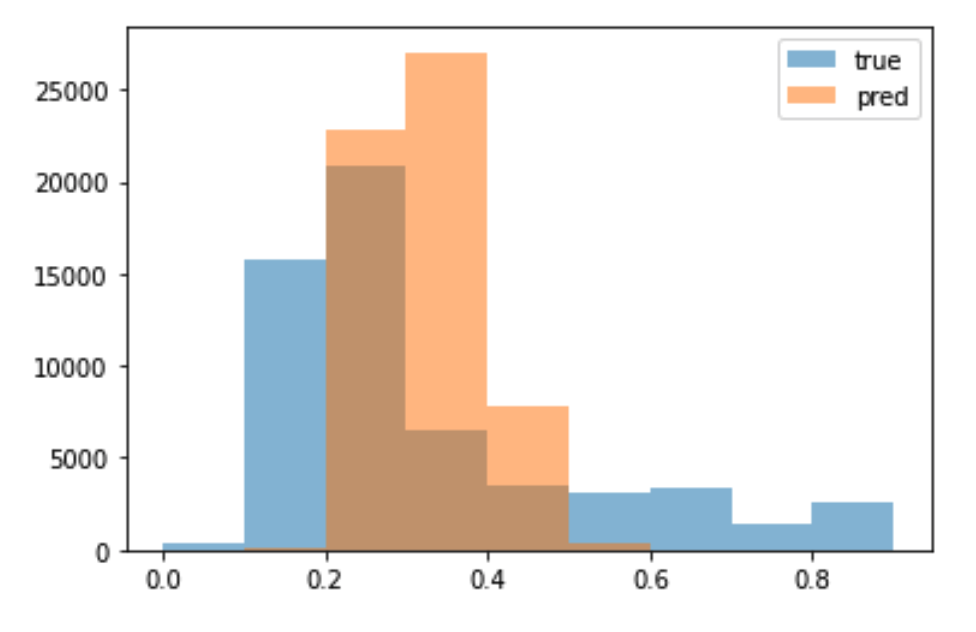}
    \caption{Histogram of mantissas for the 58K sentences in FinNews dev set (true) and corresponding predictions by DExp (pred).
    See Section \ref{sec:results} for details.
    }
    \label{fig:man}
    \vspace{-20pt}
\end{figure}

\textbf{Comparing Mantissas} (Figure \ref{fig:man}). 
To study why Vocabulary change is nearly as good as Regression, we dig deeper into the only component that differentiates our proposed Vocab-AM/GM models from the state-of-the-art DExp: mantissas. 
We plot the mantissas from DExp's predictions against the ground truth for FinNews dev set. 
We find that in the naturally occurring datasets, the leading digit of numbers is likely to be small (Benford's Law) and the mantissa peaks around 2, owing to the frequent mentions of years since 2000 (Recency Bias). This simple distribution of numbers in the real world helps a static Vocab-AM/GM model perform at par with state-of-the-art without making any architectural changes to the underlying language model.

\vspace{-5pt}
\section{Related work}
\vspace{-5pt}
\label{sec:related}

We restrict our analysis to the task of \emph{approximately} decoding numbers in MNP setting, which requires different methods and metrics from the tasks that instead evaluate their \emph{exact} arithmetic skills \citep{thawani-number-survey}.
The method we highlight in this paper \ie change of vocabulary to tokenize numbers on a log-scaled number line, has been previously used in different settings.  
Others have shown the benefits of using such exponent embeddings as \emph{number encoders} for language models, whether it be for the task of masked number prediction \citep{berg-kirkpatrick-spokoyny-2020-empirical} or masked word prediction \citep{thawani-etal-2021-numeracy}. Our work extends these results with further evidence of the representational power gained by simply tokenizing numbers on the number line.

Our simple intervention to improve \emph{approximate} numeracy in LMs is also related to other work \cite{Chen2022ProgramOT} which aims to improve \emph{exact} numeracy of LMs without any architecture change.

\vspace{-5pt}
\section{Conclusion}
\vspace{-5pt}
Subword tokenization, the standard approach to representing numbers leads to inaccurate numerical understanding. 
In this work, we analyze number representation approaches that make notational (\eg scientific vs. decimal), vocabulary (\ie tokenizing on the number line), and architectural changes (\ie regressing to the number). 
We find that tokenization on the number line achieves near or better than state-of-the-art 
results 
while requiring minimal intervention to the language model.

\emph{This is a negative insight against recent results in the community which suggest that language models must be architecturally modified to gain numeracy}. 
It will allow language models to conveniently improve their numeracy, including cases where users may not have access to the model's architecture and are only provided a typical finetuning regime with small changes to the tokenizer's vocabulary.
Finally, we find similar trends in the challenging setting of numerical fact estimation for solving Fermi Problems -- indicating that vocabulary-change is sufficient to represent approximate numbers effectively with minimal effort.

\section{Acknowledgements}
This work was funded by the Defense Advanced Research Projects Agency with award N660011924033.
We would like to thank Peter Clark (AI2) for insightful discussions on the project, and the
anonymous reviewers at EACL 2023 and Negative Insights workshop for helping us refine earlier versions of this paper. 

\section{Ethics and Limitations}
\label{limit}
Our findings and recommendations may not apply beyond the English language and the Hindu-Arabic Numeral system, which are by no means the only language / number systems in use today. We encourage follow-up work to take other systems into consideration, on the lines of \citet{johnson-etal-2020-probing} and \citet{nefedov2020dataset}.
Our recommended method of tokenizing on the number line is lossy by design. It collapses several numbers into large discrete bins, and is unlikely to be suitable for exact numeracy as is required for, say, math word problems. We note that an ideal number representation should capture both approximate and exact numeracy.


\bibliography{custom}
\bibliographystyle{acl_natbib}

\section{Appendix}

\subsection{Implementation Details}
\label{appendix}
Each of our experiments took a few hours on NVIDIA Quadro RTX 8000 GPU (one per experiment).
We report results on the same random seed across models. We were able to reproduce DExp result scores exactly up to 1 decimal place. For legibility in result tables, we skip variance estimates (bootstrapped over 10 samples, each of size 75\% of the test set) in Table \ref{tab:bert} -- they range from 1e-7 to 1e-5. Note that we only compare number decoders and not the encoders -- therefore, when numbers are present in the input, standard encoding schemes are used. For approaches with changes to vocabulary and architecture, we follow \citet{berg-kirkpatrick-spokoyny-2020-empirical} and use exponent embeddings to encode numbers (with no shared parameters with the decoder's tokens) and for approaches with notation changes, we use subword tokenization. 

The key contribution of this work is to highlight the possibility of achieving near state-of-the-art results from \citet{berg-kirkpatrick-spokoyny-2020-empirical} with a much simpler method. Thus, we used the same hyperparameters and extend their code\footnote{\url{https://github.com/dspoka/mnm}} for most of our experiments.
Please refer to Section 3 in their paper for dataset details.

With scientific notation, a previous approach NumBERT \citep{zhang2020scale} denotes $329$ as $329$ \texttt{[EXP]} $2$. However, we find that representing the same instead as $3 x 29$ where `x' is the common English alphabet, works better in practice. 

\subsection{Example predictions}
\label{egs}
\begin{table*}[]
    \centering
    \begin{tabular}{lcc}
    \toprule
    \textbf{Input} & \multicolumn{1}{l}{FY2018 Earnings per share view} & \multicolumn{1}{l}{Daniels maintains Cohen paid her \$130000 via essential} \\
    & \multicolumn{1}{l}{\$ \textbf{[MASK]} , revenue view \dots} & \multicolumn{1}{l}{consultants to hush up a \textbf{[MASK]} s. encounter with Trump.} \\
   \midrule
    \textbf{True} & 1.63 & 2006 \\
    \midrule
    \textbf{Sub} & 1000000 & 1 \\
    \textbf{DExp} & 2.695 & 2792.66  \\
    \textbf{Ours} & 1-10 & 1k-10k \\
    \bottomrule
    \end{tabular}
    \caption{Example predictions from FinNews dev set. Ours (Vocab-GM) and DExp estimate numbers in the same order of magnitude as ground truth; but the subword baseline (Sub) is far off.}
    \label{tab:egs}
\end{table*}

Table \ref{tab:egs} shows some representative examples from FinNews dataset where the Subword baseline's estimate is far off from the ground truth, whereas predictions of both DExp and Vocab-GM are within the correct order-of-magnitude.

\subsection{Variable Length Binning}
\label{range}

Motivated by the success of frequency-based surface-level vocabulary, we further experiment with an extension of the vocabulary change. Instead of collapsing numbers into order-of-magnitude or exponent bins which are equally spaced on the log scale, we find bins such that their overall frequencies in a corpus are more uniform. By arranging all numbers from the FinNews corpus in ascending order and dividing them into equal sized (by frequency) bins, we get the following variable length vocabulary: $1, 2, 3, 4, 6, 10, 14, 21, 30, 31, 70, 415, 2011, 2017,$ $2018, 5131, 30207, 252178, 1700000, 30000000,$ $1152337024$. With these 21 bins\footnote{We manually tune this hyperparameter so as to obtain a near-uniform distribution of number occurrences.}, we retrain the Vocab-AM method and compare with our earlier static bins which corresponded to powers of 10: $1, 10, 100,\dots$.

Table \ref{range} shows the results on both FinNews and FinNews-\$ datasets. We observe that this vocabulary, despite having a more uniform distribution of numbers, does not do any better than the original naive method (except on LogMAE over the FinNews dataset). We note this as further evidence of the robustness of merely tokenizing on the number line. If variable sized bins were crucial for strong performance, practitioners may have had to relearn the model's numeric vocabularies based on different datasets and corpus frequencies. On the other hand, the order-of-magnitude-10 vocabulary is a simple, intuitive and robust method that competes with performance of state-of-the-art architectural-change number decoders.

\begin{table}[]
    \centering
    \begin{tabular}{l|cc|cc}
    \toprule
    & \multicolumn{2}{c|}{FinNews} & \multicolumn{2}{c}{FinNews-\$} \\
    \textbf{Metrics} & E-Acc $\uparrow$  & \multicolumn{1}{c|}{LogMAE$\downarrow$} & E-Acc$\uparrow$ & LogMAE$\downarrow$ \\
    \midrule
    Vocab-AM & \underline{74.40} & 0.65 & \underline{57.14} & 0.93 \\ 
    Vocab-GM & 73.70 & \underline{0.60} & 56.99 & \underline{0.92} \\
    \textbf{DExp-21} & 72.2 & 0.51 & 47.6 & 1.04 \\
    DExp & \textbf{74.56} & \textbf{0.50} & \textbf{57.50} & \textbf{0.89} \\
    \bottomrule
    \end{tabular}
    \caption{Comparing variable sized numeric vocabulary (Vocab-21) with static variants and architecture change (DExp) shows no gains, except in LogMAE over Financial News dataset. See \S \ref{range} for details.}
    \label{tab:range}
\end{table}

\subsection{Neuron Probing}
\label{neuron}

In this subsection, we further probe how numeracy is stored in the feed forward layers of language models. Previous work along these lines \citep{geva-etal-2021-transformer} have shown promise in interpreting the knowledge stored in language models by finding individual neurons in feed forward layers that are triggered by specific patterns of input. We apply this analysis to find some such neurons, if any, which can effectively and efficiently capture the magnitude of a masked number.

\begin{figure*}
    \centering
    \includegraphics[width=0.95\textwidth]{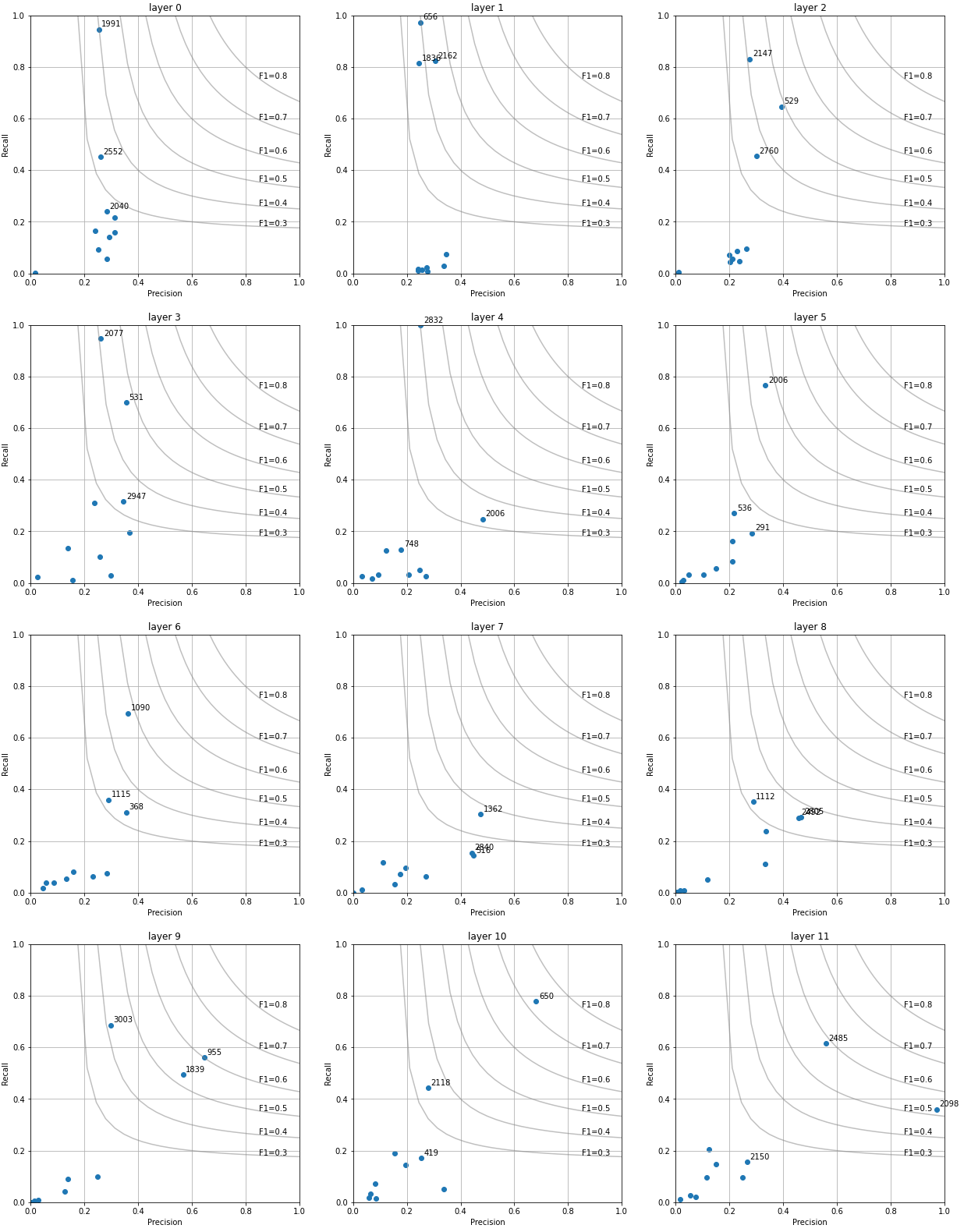}
   \caption{Precision Recall curve for the state-of-the-art (architecture-change) DExp model on the task of predicting masked numbers has an exponent of 3, \ie it is between 1000 and 10,000. See Section \ref{neuron} for details.}
    \label{fig:pr}
\end{figure*}

Figure \ref{fig:pr} shows the Precision-Recall curves for the state-of-the-art DExp model on the task of predicting masked numbers has an exponent of 3, \ie it is between 1000 and 10,000. We say a neuron has been triggered if it is among the top 50 activated ones (out of 3072) in that layer for the input mask token. Recall is then defined as the fraction of times when this neuron was triggered for all masked numbers with an exponent of 3. Precision is defined as the fraction of times when the exponent was 3 for all the times that the specific neuron was triggered. We find that some individual neurons, such as the 650th neuron in the 10th layer of finetuned DExp has a very high precision and recall. It alone can predict whether the order of magnitude is 3, with an F1 score of above 0.7. 

The presence of such precise individual neurons that capture order-of-magnitude numeracy in DExp model further suggests why tokenizing the number line on the log scale is a naturally suited number representation.
This analysis shows promise in interpreting results of number representations in language models and possibly even causing interventions to update its beliefs \citep{dai-etal-2022-knowledge}.

\end{document}